\documentclass[runningheads]{llncs}
\usepackage[T1]{fontenc}
\usepackage{graphicx}
\usepackage{amssymb,amsmath}

\def\LGEM{LGEM\textsuperscript{+}}
\def\Scv{\textit{Saccharomyces~cerevisiae}}
\def\scv{\textit{S.~cerevisiae}}

\begin{document}

\title{The Use of AI-Robotic Systems for Scientific Discovery}

\author{
    Alexander H. Gower\inst{1}\orcidID{0000-0002-8358-0842} \and
    Konstantin Korovin\inst{2}\orcidID{0000-0002-0740-621X} \and
    Daniel Brunns\aa ker\inst{1}\orcidID{0000-0002-5167-0536} \and
    Filip Kronstr\"{o}m\inst{1}\orcidID{0000-0002-3011-5541} \and
    Gabriel K. Reder\inst{5}\orcidID{0000-0001-8918-0789} \and
    Ievgeniia A. Tiukova\inst{1,3}\orcidID{0000-0002-0408-3515} \and
    Ronald S. Reiserer\inst{4}\orcidID{0000-0002-3786-7893} \and
    John P. Wikswo\inst{4}\orcidID{0000-0003-2790-1530} \and
    Ross D. King\inst{1,5}\orcidID{0000-0001-7208-4387}
}

\authorrunning{A. H. Gower et al.}

\institute{Chalmers University of Technology, Gothenburg, Sweden
    \email{\{gower,danbru,filipkro,tiukova,rossk\}@chalmers.se} \\
    \and
    The University of Manchester, Manchester, United Kingdom
    \email{Konstantin.Korovin@manchester.ac.uk} \\
    \and
    KTH Royal Institute of Technology, Stockholm, Sweden \\
    \and
    Vanderbilt University, Nashville, TN, USA \\
    \email{\{ron.reiserer,john.p.wikswo\}@vanderbilt.edu} \and
    University of Cambridge, Cambridge, United Kingdom \\
\email{gr513@cam.ac.uk}}

\maketitle

\begin{abstract}
    The process of developing theories and models, and testing them with
    experiments is fundamental to the scientific method.
    \nobreak{Automating} the
    entire scientific method then requires not only automation of the
    induction of theories from data, but also experimentation from design
    to implementation. This is the idea behind a robot scientist---a
    coupled system of AI and laboratory robotics that has agency to
    test hypotheses with real-world experiments.

    In this chapter, we explore some of the fundamentals of robot scientists
    in the philosophy of science. We also map the activities of a robot
    scientist to machine learning paradigms, and argue that the scientific
    method shares an analogy with active learning.

    We relate these general guiding principles for designing robot
    scientists to examples from the domain of system
    biology: Adam, Eve, and Genesis.
    We present a case study of Genesis, a next-generation
    robot scientist designed for research in systems biology,
    comprising a micro-fluidic system with 1000 computer-controlled
    micro-bioreactors. We discuss \LGEM{}, a logic-based model that is
    used in Genesis, and related them to discussed general principles.

\end{abstract}

\section{Introduction}\label{introduction}

In the past two decades, the use of AI-robotic systems in scientific
research has been demonstrated not only possible, but fruitful.
Several projects---from Adam~\cite{kingAutomationScience2009} and Eve
\cite{williamsCheaperFasterDrug2015} to the Robot Chemist
\cite{burgerMobileRoboticChemist2020}---have proved the value of coupling
AI software agents with experimental platforms to give them real-world agency.
A commonly used term to describe such systems is \textit{robot
scientist}, defined in King et al.~\cite{kingAutomationScience2009} as:

\begin{quote}
    ``a physically implemented laboratory automation system that
    exploits techniques from the field of artificial intelligence
    to execute cycles of scientific experimentation.''
\end{quote}

\noindent Robot scientists have the potential to broaden and deepen the
capability of the scientific community, enabling high-throughput
science and new modes of research, as well as helping to address
problems with reproducibility, and human resource bottlenecks
\cite{zenilFutureFundamentalScience2023}. Research toward the goal of
productive robot scientists is necessarily a multi-disciplinary
endeavour. Fields that contribute include: artificial intelligence,
robotics, nanotechnology and materials science. For robot scientist
projects to be a success, researchers need a shared understanding of
the process they are working to automate: the scientific method. This
chapter aims to provide researchers with knowledge and tools helpful in
analysing and designing robot scientists.

Section~\ref{the-philosophy-of-robot-scientists} first defines some
of the core concepts in the philosophical ideas behind robot
scientists, beginning with theories and model---the fundamental
entities in scientific research. Then, methods of inference are
introduced and how these should be considered when designing robot
scientists. We explore the concept of parsimony as it relates to the
scientific method, something which is learned through scientific
education in humans but requires explicit care when creating robot
scientists. The section concludes with a description of the different
scientific values used to assess the value of a theory or model.
Section~\ref{scientific-discovery-as-machine-learning} aims to
identify which machine learning paradigm is most apt for scientific
discovery. Ultimately, we conclude that scientific discovery shares
an analogy with active learning, and that this, rather than
reinforcement learning, is the most useful paradigm to adopt when
designing or analysing robot scientists.
Section~\ref{biological-systems-are-good-target-for-scientific-discovery-automation}
gives an overview of systems biology, the domain in which the next
generation robot scientist Genesis is applied. We argue that because
biological systems are ``complex systems'' as described by Simon
\cite{simonArchitectureComplexity1962}, they are excellent targets
for study by robot scientists with their superhuman abilities in
reasoning and precision. In Section~\ref{case-study-genesis-and-lgem}
we present a case study of Genesis, and one computational model
developed with the aim of automated scientific discovery.

\section{The Philosophy of Robot
Scientists}\label{the-philosophy-of-robot-scientists}

Part of the motivation for building robot scientists is to understand
more about the nature of science by building a system that can replicate
the scientific process~\cite{kingAutomatingSciencesPhilosophical2018}.

The goal of science is to develop theories that explain and predict
phenomena in the real world. To develop theories, science uses models,
which are representations of theories in some localised context.
Models are a surrogate for the system being studied (object system).
This means they have characteristics or behaviour sufficiently similar
to that of the object system to allow indirect study of the object
system by studying its surrogate. Models are particularly useful when
direct study of a system is impossible, impractical or undesirable.

\begin{samepage}
    Two examples of models in biology that illustrate the diversity of
    desirable properties of a model are:
    \begin{enumerate}
            \def\labelenumi{\arabic{enumi}.}
        \item
            an illustrated diagram of the cross-section of a cell; and
        \item
            a metabolic network model (MNM) representing the rates of
            biochemical
            reactions and chemical compound abundances using a system
            of ordinary
            differential equations (ODEs) with independent variable \(t\), time.
    \end{enumerate}
\end{samepage}

In both of these cases, the object system is the same: a cell. However
they have quite different qualities. Model~1 would be well-suited to
teaching high school students how cells of the yeast
\Scv{} function. However, Model~2 is capable of
quantitative predictions, enabling direct comparison with quantitative
experimental data.

Models with deductive capacity using mathematics, such as Model~2, are
useful for all forms of scientific discovery, but particularly when
automated. The scientific discovery problem becomes, as defined in
Flach~\cite{flach2006abduction}:

\begin{quote}
    ``an incremental process of refinement {[}of the model{]} strongly
    guided by the empirical observations.''
\end{quote}

\noindent Methods of scientific enquiry rely on: constructing a good starting
model; inferring changes to the model; techniques to reason about
which models are better; and the collection of relevant, high-quality
empirical data to drive the process. Some of these activities are
domain-specific, but there are elements that are common among
sciences, particularly inference
(Section~\ref{core-aspects-of-scientific-method})
and model evaluation
(Sections~\ref{core-aspects-of-scientific-method}~and~\ref{comparing-scientific-models}).

The purpose of a robot scientist is to provide software and hardware that
together can achieve each of these activities, and join them
together in a ``closed-loop'' form of enquiry without human intervention
\cite{kingAutomationScience2009}. Through the process of designing a
system capable of independent scientific inquiry, we seek insight into
the scientific method itself, as well as the system subject to enquiry.

\subsection{Elements of scientific
method}\label{core-aspects-of-scientific-method}

There are three components of the scientific method considered here:
logical inference, statistical inference, and parsimony. A
history and detailed treatment of each of these components is covered in
Gauch~\cite{gauchjrScientificMethodBrief2012}; here we provide a brief
introduction and highlight how these concepts can be used to analyse
robot scientists and how they shape decisions during their design.

\subsubsection{Logical inference}\label{logical-inference}

Logics are mathematical languages that relate premises and
conclusions. Logics are used in science to represent facts
(observations of the real world) and laws (parts of theories or
models)
\cite{carnapIntroductionPhilosophyScience1974,gauchjrScientificMethodBrief2012}.

There are three basic forms of logical inference: deduction,
induction and abduction. With deduction, conclusions are derived from
premises and laws. A valid deductive argument is one that guarantees
the truth of its conclusions given the truth of its premises.
Deduction is what enables deterministic simulations, and how we
reason about a hypothesis in the scientific method.

Induction and abduction cannot provide guarantees of truth. Both seek
explanations for a set of observable facts. Induction seeks laws that
can explain a general case, whereas abduction seeks facts that can
explain specific cases.

A simple example of induction is: having observed 10 yeast
cultures that thrived in a sugar solution, and another 10 that died
in plain water, one \textit{induces the law} that yeast need sugar to
thrive. While this law fits the observations, it turns out to be
false in general; some yeasts, including \scv{} can grow on ethanol.

An example of abduction is: given the law induced above,
and an observation that a yeast culture is thriving in a liquid of
unknown composition, one \textit{abduces the fact} that there is
sugar in the liquid.

Robot scientists require concretely defined induction or abduction
problems to make contributions to scientific knowledge. In the case
of the first robot scientist, Adam, the scientific problem was to
identify links between \scv{} genes and biochemical reactions
\cite{sparkesRobotScientistsAutonomous2010}. By
using Enzyme Commission (E.C.) numbers to refer to classes of enzymes
which catalyzed a given reaction, the problem was transformed to one
of abduction; Adam was to find facts of a given form.

\begin{quote}
    ``A single hypothesis is the mapping of one \scv{} ORF {[}gene{]}
    to one E.C. class - e.g. YER152C $\longrightarrow$ 2.6.1.39.''
\end{quote}

\noindent The discovery algorithm therefore had a well-defined output
form. This was possible because the laws applying to E.C. classes
were defined in Adam's logical model. In other cases, when the laws
are unknown or not well enough defined, induction will be necessary,
and to make these problems concrete one should either use techniques
for induction that provide explanations---e.g. inductive logic
programming---or ones that do not---e.g. learning neural network models.

Both induction and abduction deal in uncertainty, and therefore need
probabilistic and statistical inference.

\subsubsection{Statistical inference and
probability}\label{statistical-inference-and-probability}

Certain laws in science may only be expressed to a degree of
certainty, and empirical data are subject to random effects. These
situations require statements that can deal with uncertainty, or
probability, of which two types are defined by Carnap
\cite{carnapIntroductionPhilosophyScience1974}.
\textit{Statistical~probability} is a frequentist idea about the
relative frequencies of mass events. The defining characteristic of
statements of statistical probability is that they cannot be decided
by logic, but rest on empirical observations.
\textit{Logical~probability} on the other hand is the probability on
a logical relation between two propositions. Gauch prefers to
describe the two types of probability as being about events and
beliefs respectively~\cite{gauchjrScientificMethodBrief2012}.

Statistics can be used to obtain a model by reasoning about observations. In the
case of abduction, one is reasoning about facts; with induction, laws.
The statistical element of this reasoning represents the uncertainty
around the model, which can be viewed either as a belief in the law, or
perhaps more commonly in empirical science, the relative frequencies of
events. Either way, these processes are crucial to the scientific method
because they allow us to form and improve upon models.

Statistical reasoning can be done when evaluating
hypotheses against empirical data, or during the hypothesis generation stage.
Adam, for example, used statistical measures of gene sequence
similarity (PSI-BLAST, FASTA) to identify candidate genes in \scv{} from genes
in other organisms~\cite{sparkesRobotScientistsAutonomous2010}. And
when designing the most recent discovery framework for the robot
scientist Eve, Brunnsåker et al. used inductive logic programming to
identify candidate phenotypes~\cite{brunnsakerAgenticAIIntegrated2025}.

\subsubsection{Parsimony}\label{parsimony}

Parsimony is a term that refers to two different concepts.
\textit{Epistemological parsimony} is the concept that when choosing
between theories that fit the data equally well, the theory that is
simplest is preferable, sometimes referred to as ``Ockham's razor''.
\textit{Ontological parsimony} is the idea that nature itself prefers
simplicity. Both are both absolutely essential for science, and it is
easy to just take parsimony as a common sense notion. However, a
proper treatment of the motivations for adopting parsimony as a
guiding principle is warranted when it comes to systems biology and
the automation of scientific discovery.

Concrete examples of ontological parsimony include: taking the assumption
that there exist common properties between individuals of the same
species. This could be a chemical species, that all glucose molecules
react with water molecules in the same ways. Or it could be a species of
organism, implying that the same computational model can be applied to two
individual colonies of \emph{S. cerevisiae}.

Arguments for ontological parsimony suffer from many
counter-examples. For example the yeast genome went through a duplication
during its evolution resulting in numerous genes with overlapping or
identical function~\cite{kellisProofEvolutionaryAnalysis2004}. However
all scientific argument must adopt some version of ontological parsimony,
as it is the basis for generalisation of theories.
Ideas of ontological parsimony are so central to scientific enquiry
that researchers may not consider them---they become necessary implicit
biases.

One power of appealing to ontological parsimony is that it is the basis
of factorial experimentation, described in
\cite{carnapIntroductionPhilosophyScience1974} as a two-step process:
firstly, identify relevant factors for the phenomenon to be studied; then
design experiments holding certain factors constant and varying over
others. Determining relevant factors means stating that certain factors
are irrelevant, i.e. that they are not expected to affect the outcome of the
experiment. Deciding which factors to include is not an easy process.
Designing experiments in this way allows the controlled study of
phenomena to test hypotheses relating to a restricted
subset of factors. Factorial design of experiments requires an appeal to
ontological parsimony, which we can informally express as:

\begin{quote}
    \begin{minipage}{\dimexpr\linewidth-3em}
        empirical data from experiments of the same class are expected
        to exhibit only random variation, where the class of
        experiments depends only on the variable factors.
    \end{minipage}
    \hfill
    (A)
\end{quote}

\noindent This
assumption allows the inference of empirical laws about the phenomenon.

In practice one cannot usually keep constant all the factors one would
like, so the class of experiments does not only depend on relevant
variables. The nature of a given experimental protocol will introduce
systematic errors: variations in the empirical data not arising from the
experimental variables or random noise. Many methods exist to mitigate
and model systematic errors, using randomisation techniques, systematic
design, and statistics. Randomisation, or
systematic designs like Latin squares, can be impractical when using
robot scientists due to the limitations of the automation hardware. One
example from biology is that liquid handling procedures for Latin
square designs are impossible on certain liquid handling robots, and
can increase
procedure times by orders of magnitude on those with such a capability.

On the other hand, robot scientists have advantages when it comes to
relying on and examining Statement~(A). Firstly, that robots are
capable of performing repeated tasks with a much higher accuracy and
precision than human counterparts, as is shown by examples of adoption of
laboratory automation in biology~\cite{hollandAutomationLifeScience2020},
physics~\cite{roccaprioreAutomatedExperiment4DSTEM2022}, and chemistry
\cite{shiAutomatedExperimentationPowers2021}. Secondly, that the
validity of Statement~(A) can be evaluated by recording more data about the
execution of experiments than usually recorded when humans complete
experiments. This is a ``natural by-product''
\cite{kingAutomationScience2009}, as robots and software frequently have
automatic logging capability. Finally, that those who design robot
scientists encode the
experimental protocol forces them to specify which experimental
factors will be constant, or that if they are not constant they are
judged irrelevant, and therefore they do not break the validity of
statement above by creating a new class of experiments (for example, the
    use of different individual glass flasks of the same brand, model, and
age).

Epistemological parsimony is a concept much more familiar to human
scientists, or at least one which is applied more explicitly. It has been
formalised in information theory as the minimum message length (MML):
that the hypothesis that best explains the data is that which
minimises the total information. The total information has two
components, the \textit{a priori} information contained in the
hypothesis, and the information of the data given the hypothesis\footnote{
    Combine Shannon's formula for the information (\(I\)) of an event
    (\(E\)), given by \({I(E)=-log_2(P(E))}\), with Bayes' theorem, \({P(E_1\cap
    E_2)=P(E_1)P(E_2|E_1)}\). MML states that the hypothesis \(H\) that best
    explains data \(D\)---in other words it maximises \({P(H\cap D)}\)---is
    that which minimises the information (message length): \({I(H\cap
    D)=I(H)+I(D|H)}\).
}. There is a trade-off between these two quantities. One
can make a hypothesis so specific as to explain all the data, meaning
the information content of the data given the hypothesis is zero. Or
one can have no hypothesis at all. In practice, MML ensures that
information is only added to a
hypothesis if this information explains the data, which is
the principle of Ockham's razor
\cite{allisonCodingOckhamsRazor2018,barnardStudiesHistoryProbability1958,shannonMathematicalTheoryCommunication1948}.

Epistemological parsimony is the basis for several fundamental concepts
in the modern scientific method. One example is the idea of a null
hypothesis---a statement that any variance in
empirical data can be explained by the extant model, to a degree of
certainty due to random noise. To reject the null hypothesis is to accede
a more complex model, and is only done if the current
model is insufficient to explain the empirical data.

For robot scientists, epistemological parsimony should also be
applied in the choice of model as it relates to the experimental
hardware. More complex models that suggest hypotheses outside of set
of possible experiments result in inaction. Adam was able to achieve
autonomous discovery in part because of the parsimony of the logical
theory used to generate hypotheses and evaluate them. There were many
more facts and relations that could have been included in the theory,
but by omitting these and building a theory that was focused on the
scientific discovery task, the theory was tractable and resulted only
in hypotheses that were testable by the experimental apparatus
available to the robot.

\subsection{Comparing scientific models}\label{comparing-scientific-models}

The relative merit of competing scientific models is not a trivial
assessment. This problem is referred to in the philosophy of science as
the problem of \emph{theory choice}. For a robot scientist to be
effective, its design must incorporate values and an evaluation procedure,
otherwise the scientific discovery process will depend on human
evaluation.

In discussing whether scientists follow philosophical virtues in their
methods, Schindler~\cite{schindlerTheoreticalVirtuesScientists2022}
presents six virtues following Kuhn
\cite{kuhnEssentialTensionSelected1977}.

\begin{itemize}
    \item
        \textbf{Internal consistency} is defined as the absence of
        contradictions within a theory. This can be extended to
        include the various contexts in which the theory is applied;
        in biology this could mean growth of \emph{S. cerevisiae} in
        different conditions.
    \item
        \textbf{External consistency} is the absence of
        contradictions with other scientific theories. For example,
        that a model of the biochemistry of yeast is
        thermodynamically consistent.
    \item
        \textbf{Empirical accuracy}, otherwise referred to as
        predictive power, is the degree to which deductions from
        the theory match observations. For example, predictions of
        growth rates for colonies of yeast.
    \item
        \textbf{Scope}, otherwise referred to as unifying power, is
        the quality that a theory explains concepts relevant to
        different phenomena. An example in biology is the existence of a
        unifying theory of genetics for DNA-based life, rather than
        separate theories for different species or kingdoms.
    \item
        \textbf{Simplicity}, which comes in various forms, depending
        on the context. Kuhn relates this to tractability using the
        example of Ptolemy's and Copernicus' theory of astronomy and
        the number of calculations needed for prediction being
        equivalent in both systems. In some theoretic sense,
        Copernicus' theory was simpler than Ptolemy's, having a
        simpler mathematical formulation.
    \item
        \textbf{Fruitfulness (or fertility)}, which has various
        interpretations. But can in one way be understood to be: how
        well does this theory lead to ``more science''? When combined
        with scope, fruitfulness leads to models that generalise well to new
        applications or other domains. Fruitfulness also has one
        clear implication for closed-loop discovery.
\end{itemize}

Schindler~\cite{schindlerTheoreticalVirtuesScientists2022}
found that scientists do not agree on the relative importance of these
characteristics, although there were some common views. And Kuhn
\cite{kuhnEssentialTensionSelected1977} argued
that even if there were a common order and weighting, that two individual
scientists may honestly differ in their assessment of a better theory
because of the ways they evaluate them. This presents a difficulty for
robot scientists as well. The difficulty of differences in evaluation is
common to humans, so we can accept this as part of science. The
unique difficulty for robot scientists is imbuing these values into
the software used to evaluate theories.

Models trained on human knowledge will pick up some of the implicit
biases in these data, which could be seen as a way to learn these values
implicitly. This is particularly true of using foundation models such as
large language models (LLMs, covered in
Section~\ref{scientific-discovery-as-machine-learning}) to evaluate
theories. However, this can present risks, due to the lack of knowledge
we have on their training and our inability to interrogate them.

Thankfully, the problem is not as difficult as it may initially appear,
as machine learning research is itself informed by similar values. The
task then for designing a robot scientist is to align mechanisms from AI
research with scientific values in the relevant domain. A domain-specific
cost function ensures accuracy is considered; likewise regularisation
terms cover simplicity; and there are active research areas in enforcing
external consistency on machine learning models, for example imposing
symmetry constraints from physics on to neural
networks\cite{akhound-sadeghLiePointSymmetry2023}.

\section{Scientific Discovery as Machine
Learning}\label{scientific-discovery-as-machine-learning}

Satisfied that automating scientific discovery can be considered a
machine learning problem, the question remains: how best should
methods from the field of machine learning be applied to scientific
discovery? This section covers how to view the components of
scientific discovery, as presented in
Section~\ref{the-philosophy-of-robot-scientists}, as components of
machine learning techniques, and investigates which of the machine
learning paradigms are the most appropriate for application to robot scientists.

Common to all machine learning techniques is that given some data to
train with, the goal is to learn a function that will perform in a desirable
manner, and generalise beyond the given data.

In terms of the logical reasoning components of scientific discovery,
forming a model (\(\hat{h}\)) from data is an induction problem.
Machine learning seeks to form a model using information from a
relevant signal to assess performance against a goal; this could mean
defining a loss function for an optimiser, or in the case of
inductive logic programming explaining positive examples
and avoid inconsistencies with negative examples. To evaluate a
candidate model, \(\hat{h}\), predictions or
consequences are obtained from the model (deduction) and then evaluated
against the signal, often using a form of statistical reasoning. And
finally, it is usually the case that there are many possible models
that may have similar performance against the goal. In which case,
most machine learning techniques appeal to parsimony to choose the
model which is simpler, in some defined sense. This could be the
inclusion of a regularisation term in a loss function, or selecting
the simplest logic program that covers the examples.

\begin{figure}
    \centering
    \includegraphics[width=\textwidth]{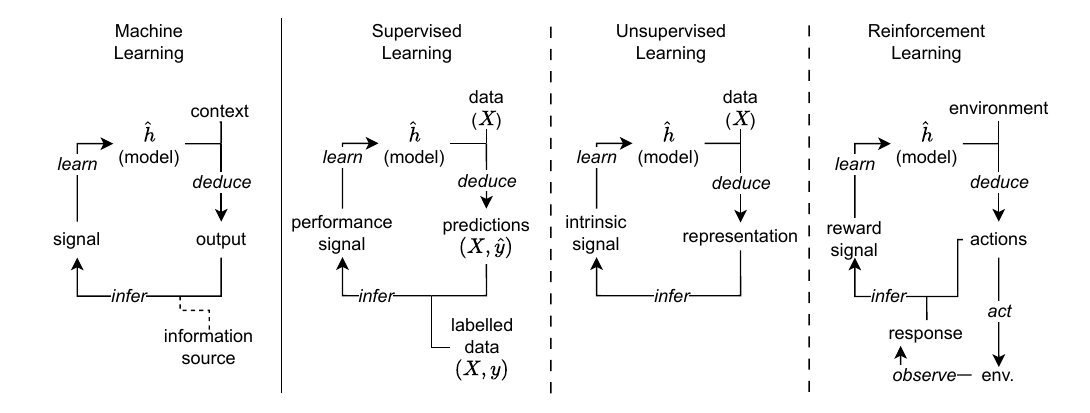}
    \caption{Flowcharts representing high level design for: generic
        machine learning; supervised learning;
        unsupervised learning; and reinforcement learning. Unitalicised
        text represents inputs and outputs; italicised labels for
    connectors represent processes.}\label{fig:ml-paradigms}
\end{figure}

Machine learning techniques can be placed into three broad categories:
supervised learning, unsupervised learning, and reinforcement learning
\cite{sarkerMachineLearningAlgorithms2021}; these paradigms are
described below and in Figure~\ref{fig:ml-paradigms}. Semi-supervised
learning is
sometimes also included as a fourth category, being a hybrid of
supervised and unsupervised learning.

\begin{itemize}
    \item \textbf{Supervised learning} covers machine learning
        techniques that work from labelled data. These are
        input-output pairs, and the machine learning task is to learn
        a function that maps input to output.
    \item \textbf{Unsupervised learning} techniques seek to find
        structure and extract information from unlabelled data. The
        signal received is an intrinsic objective measure, for
        example data likelihood.
    \item \textbf{Reinforcement learning} techniques require feedback
        from the environment in which the learning agent is embedded.
        This feedback is used to evaluate actions taken by the agent,
        which information is used to optimise a strategy for the agent.
\end{itemize}

\subsection{Reinforcement learning is an unsatisfactory paradigm for robot
scientists}\label{reinforcement-learning-is-an-unsatisfactory-paradigm-for-robot-scientists}

Recalling the definition of a robot scientist given in the introduction,
the fact that there is an agent (the laboratory automation system)
embedded in an environment (the physical surroundings of the lab) lends
us to consider that discovery using a robot scientist is a reinforcement
learning problem.

Analysing the system from the reinforcement learning perspective we must
identify: (A) the agent; (B) the environment in which the agent is
embedded; and (C) the reward function that evaluates the
agent's actions in its environment.

For pairs (A, B) there are several choices. Here we consider a
few.

\begin{enumerate} \def\labelenumi{\arabic{enumi}.}
    \item (laboratory robotic system, physical lab surroundings)
    \item (experiment selector, experimental design space)
    \item (model improvement algorithm, model space)
\end{enumerate}

In the first pair, the reward function could include measures of how
closely the robot followed the protocol and whether it dropped equipment
or created hazards such as spills. The second pair we discussed in
Section~\ref{the-philosophy-of-robot-scientists}. For third pair, we
would have to construct a reward function composed of the values
enumerated in Section~\ref{the-philosophy-of-robot-scientists}. This is a
daunting task to try and find a single function that incorporates each of
the values. And one that goes against the advice of Kuhn who said that
such a comparison would be exceedingly difficult and subjective.

In contrast to other applications of reinforcement learning, for example
autonomous vehicles or chess engines, the goal of a robot scientist is
not in and of itself to take action. Actions taken by the agents that
compose a robot scientist are in service of the broader aim of generating
new scientific knowledge.

Typically in reinforcement learning, the action to feedback time is short
and the evaluation of the reward function is cheap. This is not generally
the case for robot scientists, as physical experiments have high cost,
and usually take a significant amount of time (hours or days).

Consequently, techniques within reinforcement learning are less likely to
be applicable to the scientific discovery aspect of the robot scientist.
We conclude that reinforcement learning is unsatisfactory as a paradigm
around which to design scientific discovery algorithms for a robot
scientist. (It may well be that reinforcement learning algorithms can be
    of great use in optimising the laboratory automation, where the goal is
to take actions in an optimal way.)

\subsection{Supervised learning is a more useful
paradigm}\label{supervised-or-semi-supervised-learning-is-a-more-appropriate-paradigm}

The crucial process machine learning is applied to in scientific
discovery is that of model improvement: given a model of an object
system, how to make changes to the model such that it is more
faithful to the object system. According to the scientific values
presented in Section~\ref{the-philosophy-of-robot-scientists}, there
are numerous ways to evaluate this. However, in Schindler
\cite{schindlerTheoreticalVirtuesScientists2022} ``accuracy'' was
consistently ranked second by scientists in order of preference, only
beaten by the ``internal consistency'' of a theory.

We argue that it is useful to consider scientific discovery as a
supervised learning problem, both in observational and controlled
experimentation. In the case of observational experiments,
Medawar's ``Baconian'' experiments
\cite{medawarAdviceYoungScientist1979}, the input-output pairs will
be a partition of the overall observational data. Equally, in
controlled experiments, Medawar's ``Galilean experiments'', the
input-output pairs will be the experimental factors and the empirical
data. When designing robot scientists, various mechanisms from the
field of supervised learning can be exploited to obtain theories
which align with the scientific values stated in
Section~\ref{the-philosophy-of-robot-scientists}, with accuracy
captured in a relevant loss function. This aligns with how previous
robot scientists have operated, which we cover in more detail in
Section~\ref{biological-systems-are-good-target-for-scientific-discovery-automation}.

\subsection{Semi-supervised learning}

Unsupervised learning techniques, such as embedding and clustering,
are used in scientific discovery applications during data
representation. For example, unsupervised techniques have been used
to cluster molecular dynamics data in materials science
\cite{kyvalaUnsupervisedIdentificationCrystal2025}, and
transcriptomic data in biology
\cite{hozumiPreprocessingSingleCell2024}. In each of these cases,
human scientists analysed the output of the unsupervised learning to
draw conclusions.

As discussed above, a supervised learning design allows robot
scientists to close the discovery loop and not require human
scientists to analyse results. However, semi-supervised
learning---integrating unsupervised
techniques into supervised learning---allows for contextual
information and the structure of
theories to be exploited in the discovery task, as well as addressing
the issue that the experimental space is sparsely annotated.

Semi-supervised techniques can exploit the background knowledge that
is available to the robot scientist to make better predictions and
learn more efficiently. Gleaves et al.
\cite{gleavesMaterialsSynthesizabilityStability2023} used a
semi-supervised framework to improve synthesisability models in
materials science, giving evidence that a semi-supervised approach to
learning could provide the most promise for scientific discovery applications.

\subsection{Active learning integrates agency into supervised
learning}\label{active-learning-integrates-agency-into-supervised-learning}

While supervised learning is the appropriate paradigm for scientific
discovery, the design of a robot scientist must integrate
agency. Active learning is a specific category of supervised learning
where the learning agent chooses the next data point (input) for
which a label (output) has not been observed. This selection policy
is not the focus of the learning, and there is no requirement that
the selection policy be learned. This differs from reinforcement
learning, where the objective is to learn a good policy.
(Reinforcement learning could be used to design this agent's policy,
    or alternatively use some pre-determined policy, or a combination of
the two approaches.)

Active learning approaches select a point in the input space that has
not been observed. Points are selected, usually either because the
model has high uncertainty around that point, or to increase the
diversity of the dataset. Uncertainty can arise from poor exploration
of, or shallow exploitation in, the neighbourhood of that point. The
agent then requests a corresponding output. In many applications of
active learning this means asking for a human or expert annotation.
In scientific discovery, this could be the case, or in the case of a
robot scientist it can use its experimental platform to perform an
experiment to get output data. We see that active learning shares an
analogy to the scientific method, and therefore is an appropriate and
useful paradigm to choose for the design and analysis of robot
scientists. Both Adam and Eve used active learning by searching the
hypothesis space and executing experiments to then improve the
scientific model. This is covered in more detail in
Section~\ref{superhuman-logic-and-probabilistic-reasoning}.

\subsection{Foundation models and their use in scientific
discovery}\label{foundation-models-and-their-use-in-scientific-discovery}

Recent developments in machine learning, driven by industrial
applications and enabled by new technologies and vast amounts of data,
have resulted in widespread adoption of foundation models. Foundation
models are defined in~\cite{bommasaniOpportunitiesRisksFoundation2022} as:

\begin{quote}
    ``any model that is trained on broad data (generally using self-supervision
    at scale) that can be adapted (e.g., fine-tuned) to a wide range of
    downstream tasks.''
\end{quote}

\noindent Within the scope of this definition, most current manifestations of
foundation models are large language models (LLMs)---e.g. BERT
\cite{devlinBERTPretrainingDeep2019}, GPT-3
\cite{brownLanguageModelsAre2020}---or large multi-modal models
(LMMs)---e.g. GPT-4~\cite{openaiGPT4TechnicalReport2024}. However,
this definition also encompasses models such as Evo
\cite{nguyenSequenceModelingDesign2024} or ESM3
\cite{hayesSimulating500Million2025}, transformer-based models trained
on genomic data and protein data respectively.

Foundation models show promise in scientific discovery applications.
Besides AlphaFold there is Coscientist, a system built around GPT-4 that
could autonomously design, plan and execute experiments in
chemistry~\cite{boikoAutonomousChemicalResearch2023}. Coscientist
exploited the general purpose nature of the foundation model to combine
information from various sources and to execute code and instructions on
machines to achieve its goal. In many applications there is some concern
for so-called ``hallucinations'' of LLMs and LMMs, i.e. claims made by
the model with little or no justification or evidence. This is not a
problem for a robot scientist provided they use it to generate hypotheses,
as the resultant hypothesis will be tested via experiment. Hallucinations
could cause problems in the experiment design phase; Coscientist
mitigated the impact of hallucinations by grounding the LLM with database
search, and ultimately evaluated the system's performance using explicit
criteria rather than use the LLM.

Hallucinations could also cause problems if the foundation model is
applied to the evaluation and assessment of competing theories. It is a
distinct possibility that a robot scientist might justify a theory choice
based on fabricated data or through faulty logical or statistical
inference. And these models' black box nature means we cannot interrogate
them about their reasoning. The best we can do currently is prompt the
model for a post-hoc rationalisation of its reasoning, and it is not at
all clear that this is of equal value
\cite{parkDiminishedDiversitythoughtStandard2024}.

These properties of LLMs and LMMs are a problem for use in
``closed-loop'' discovery in a robot scientist. Scientific models
should be interpretable and usable by other scientists, displaying
the values of fruitfulness and simplicity. Foundation models often
have hundreds of millions of parameters, and methods for interpreting
their internal reasoning require further research before these models
can be considered broadly suitable for automated discovery.

From a practical and economic perspective, LLMs and LMMs frequently
require huge resources throughout the life cycle of their development.
Much focus is rightly directed to the immense electricity demands during
training, but further resources are needed during research and
development, data collection and storage, the construction and
commissioning of hardware, and in the implementation and maintenance of
LLMs and LMMs~\cite{jiangPreventingImmenseIncrease2024}.

Because of these economic demands, foundation models are often developed
by large private enterprises rather than public science bodies or
universities. This introduces risks to any scientific project dependent
on these foundation models. Code can be closed-source, the training data
and regimes are often held as trade secrets, and the models are
provisioned on third-party hardware. Each project must weigh these risks
against the clear benefits of using foundation models in scientific
discovery work.

Having covered some important areas of theory behind computational
scientific discovery and robot scientists in a domain-agnostic manner, we
proceed to examine applications in a specific domain, biology.

\section{Biological Systems are a Good Target for Scientific Discovery
Automation}\label{biological-systems-are-good-target-for-scientific-discovery-automation}

Of the scientific challenges in this 21st century, understanding the biology
of eukaryotic organisms ranks among the most consequential. Despite
fantastic advances during the 20th century of our understanding of the
fundamental components and processes in biology, and in the application
of this knowledge to medicine, engineering, agriculture, etc., we are
still some way off an accurate predictive model of the physiology of one
organism, let alone a system of broadly applicable theories, such as
those developed for physics.

Part of the reason why progress in biology is limited by today's
scientific methods is the diversity and complexity of the systems.
Hundreds of research hours can be spent in the study of one particular
gene, yet the limits of human capability and of course the economic
resource available to the researcher will hamper progression to a
complete understanding of the gene and its roles. Scientific discovery
automation has therefore great potential in biology. This is
particularly the case when adopting the systems biology paradigm.

\begin{samepage}
    The cellular physiology of eukaryotes is a complex system, in the spirit
    of the definition given by Simon~\cite{simonArchitectureComplexity1962}
    that:

    \begin{quote}
        ``given the properties of the parts {[}of the system{]} and the laws of
        their interaction, it is not a trivial matter to infer the properties of
        the whole.''
    \end{quote}

    \noindent A reductionist approach to biology (breaking down a system
        and studying
    its components) has resulted in great advances in our understanding of
    the fundamental ``parts and laws'', for example the discovery of the
    double-helix structure of DNA molecules, or that all known proteins are
    composed from the same set of 21 amino acids via translation from RNA.
\end{samepage}

To understand complex systems it is necessary, yet wholly
insufficient, to take a reductionist approach. Systems biology is an
integrationist approach to studying biological systems. The aim is to
understand how the parts and their interaction lead to the resultant
behaviour of the system, be that system a cell, an organ or an entire organism.

By way of two examples of robot scientists, Adam and Eve, that were
applied to the domain of biology, we will show why this is a suitable
domain for automated scientific discovery.

\subsection{Discovery through superhuman logic and probabilistic
reasoning}\label{superhuman-logic-and-probabilistic-reasoning}

The first robot scientist was Adam~\cite{kingAutomationScience2009},
and was the first machine to autonomously discover
new scientific knowledge. Adam was designed to cultivate bacteria and
yeast in batch under varying conditions and measure phenotype, in
this case the growth rates of the cultures over time. Adam used logic
programming to analyse the theory of \emph{S. cerevisiae} to identify
hypotheses, which it then evaluated using quantitative analysis of
the growth data. Adam hypothesised that three genes encoded for the
enzyme 2-aminoadipate:2-oxoglutarate aminotransferase (2A2OA),
previously an orphan enzyme. Even the reduced system that Adam
studied resulted in a vast logical theory. To form hypotheses at
scale in the manner that Adam did would be beyond human capabilities.
Biology is a good domain for robot scientists to reason over because
of the very large number of facts and entities involved.

Eve was a robot scientist designed to automate aspects of early stage
drug development~\cite{williamsCheaperFasterDrug2015}. Eve had
several modes of operation, but we focus here on the learning of
quantitative structure activity relationships (QSARs). As the name
suggests, they take as input the structure of a compound and predict
the activity on the assay for a particular disease. Eve used a
least-squares regression to learn the QSARs, which in turn were used
to guide synthesis of new compounds, and further refine the QSAR.
These steps are all dependent on large-scale statistical inference,
partly because data collected from biological systems are often
noisy. There is a degree of stochasticity to biological processes
that are hard for humans to understand intuitively, but that machines
are apt at modelling.

\section{Case Study: Genesis and \LGEM{}}\label{case-study-genesis-and-lgem}

At Chalmers University in Sweden we are building a next-generation robot
scientist ``Genesis''. Our goal is to demonstrate that the robot
scientist Genesis can investigate an important area of science a thousand
times more efficiently (in terms of cost and money) than human
scientists.

This is an extreme challenge for AI as the number of experiments to plan
and coordinate is several orders of magnitude more than the previous case
studies. Achieving this goal will involve advances in automated
hypothesis formation (how best to utilise background biological knowledge
and models in ML, etc.), automated experiment design (how best to
optimise gain of information with cost and time constraints), laboratory
robotic control, and scientific data analysis.

\begin{figure}
    \centering
    \includegraphics[width=\textwidth]{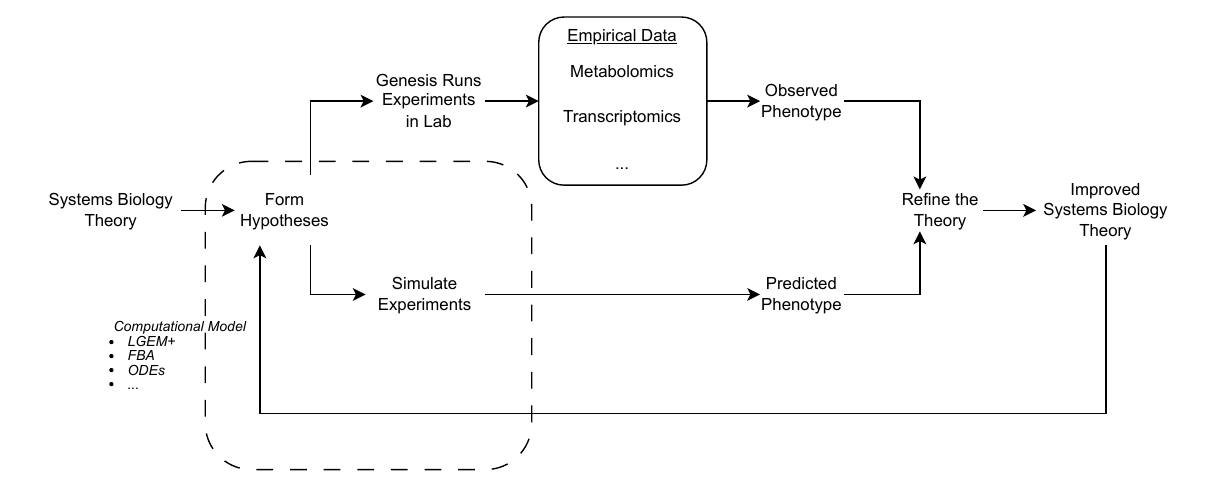}
    \caption{Flowcharts representing the scientific discovery process
        of Genesis. The robot scientist starts with a systems biology
        theory, contructed from community knowledge. After using a
        computational model (e.g. \LGEM{}) to form hypotheses, Genesis
        will design and run lab experiments using its hardware, and will
        take measurements of phenotype using automated procedures, e.g.
        for metabolomics. Simulated phenotype will be compared against
        the observed phenotype to generate information used to refine the
        theory, and the cycle will begin again with the improved
    theory.}\label{fig:genesis}
\end{figure}

The scientific discovery goal of Genesis is to develop a systems
biology model of \scv{}, that is both more detailed and more accurate
at predicting experimental results than any in existence. An outline
of Genesis' discovery process is shown in Figure~\ref{fig:genesis}.

The foundation of Genesis is a micro-fluidic system with 1000
computer-controlled micro-bioreactors (or chemostats) co-developed in
Vanderbilt University. Achieving this will be a step-change in
laboratory automation as most biological labs have fewer than 10
chemostats. These micro-bioreactors are being integrated with
ion-flow mass-spectroscopy (to measure metabolites at speed) and
RNA-seq (to measure RNA expression levels).

To design the experiments that the robot scientist conducts, and to
create and improve on the model of \emph{S. cerevisiae}, we designed
a modelling framework, which we present briefly next.

\subsection{\LGEM{}: a first-order logic
model}\label{lgem-a-first-order-logic-model}

The task of scientific discovery is described by
Langley~\cite{langleyIntegratedSystemsComputational2024} in generic
terms that given:
(a) scientific data; (b) prior knowledge about the domain; and (c) a
space of candidate categories, theories, laws, or models, scientists seek
the candidates that describe or explain the data.

In Genesis' domain, the scientific data are in the form of controlled
experiments using \scv{} and resultant empirical data. There are many types of
empirical data one could collect from such experiments. In our
discovery application we focus on metabolomics and gene expression data.

Prior knowledge on \scv{} is well-curated in genome-scale
metabolic network models (GEMs). These are community developed models
that follow a controlled vocabulary, so form a rich prior for automated
scientific discovery. We chose to express the mechanisms of the
biochemical pathways using first-order logic (FOL), an approach first
proposed in 2001~\cite{reiserDevelopingLogicalModel2001}. We use a FOL
structure that is grounded in the controlled vocabulary of the GEMs to
express knowledge about how entities are known to interact, for
example that each reaction has reactants, products, and possibly an
enzyme annotation. We call this framework \LGEM{}; below is a brief
description to illustrate the case and a more detailed explanation of the
methods is in~\cite{gowerLGEMFirstOrderLogic2023}.

\LGEM{} has five predicates: met/2, gn/1,
pro/1, enz/1, and rxn/1. These are
given specific semantic meaning, which is shown in
Table~\ref{table:predicates}. Here a cellular ``compartment'' refers to a
component of the cellular anatomy, e.g. mitochondrion, nucleus or
cytoplasm. There are seven types of clause that we included that encode
the implications needed to describe phenomena such as reaction activity
and gene expression. More detail on the specification of the logical
theories is given in~\cite{gowerLGEMFirstOrderLogic2023}.

Finally, the space of candidate theories is those able to be constructed
from the predicate symbols and constants relating to the cellular
compartments, genes, metabolites, reactions, and enzymes that could be
present in yeast.

\begin{table}[!ht]
    \caption{Predicates used in the logical theory of yeast
    metabolism, \LGEM{}.}
    \label{table:predicates}
    \centering
    \resizebox{\textwidth}{!}{
        \begin{tabular}{|l|l|l|}
            \hline
            Predicate & Arguments & Natural language interpretation \\ \hline
            met/2 & metabolite, compartment & ``Compound X is present
            in cellular \\
            & & compartment Y.'' \\ \hline
            gn/1 & gene identifier & ``Gene X is expressed.'' \\ \hline
            pro/1 & protein complex identifier & ``Protein complex X
            is available (in every \\
            & &  cellular compartment).'' \\ \hline
            enz/1 & enzyme category identifier & ``Enzyme category X
            is available.'' \\ \hline
            rxn/1 & reaction & ``There is positive flux through
            reaction X.'' \\ \hline
        \end{tabular}
    }
\end{table}

We use automated theorem provers for first-order logic (ATPs) to perform
logical inference. ATPs are software that can automatically prove or
disprove logical statements, by applying logical inference to a set
of axioms. In
comparison to previous approaches using bespoke
algorithmic methods, such as MENECO
\cite{prigentMenecoTopologyBasedGapFilling2017}, using an ATP removes a
large part of the burden of algorithm design and simulation, particularly
when it comes to abductive inference. For the reasoning tasks we use the
ATP iProver~\cite{korovinIProverInstantiationBasedTheorem2008}, which was
chosen due to its performance and scalability as well as completeness for
first-order theorem finding. We extended iProver to include abduction
inference. ATPs are designed to provide explanations, and have many tools
to simplify theories to which they are applied. All these properties make
ATPs a good choice for applications in scientific discovery.

From the logical theory, the ATP can deduce testable facts, e.g.
production of metabolites. This allows us to generate input-output
pairs that can be tested against truth data. The first truth data the
model predictions were tested against were single gene essentiality
data for \emph{S. cerevisiae}. Essential genes are those genes whose
removal from the genome leads to a loss of viability for the
organism. Single-gene essentiality was predicted for \emph{S.
cerevisiae} by providing: as input (the theory \(T\)) the yeast
genotype (including the deletion), metabolites that were present in
the growth medium, and metabolites assumed to be ubiquitous in the
cell, along with the rest of the theory containing rules for
activation of reactions and formation of enzymes; and deducing the
output (the goal \(G\)) as a binary outcome of whether every
metabolite assessed to be essential for growth was produced.

Predictions were compared against empirical data for single-gene
essentiality. The F1 score on the prediction task was 0.266, which was
state-of-the-art for a qualitative method on that task, but still quite
far away from the best quantitative models. In the case that a particular
mutant is falsely predicted essential, this shows that there is room for
improvement to the model, that the robot scientist needs to come up with
hypotheses \(H_i\) such that combined with the theory this hypothesis
entails the goal (\(T \land H_i \vDash G\)). The ATP achieves this through
reverse consequence finding, using a rearrangement of the previous
statement: \(T \land \lnot G \vdash \lnot H_i\). It is also possible
to steer iProver to find specific forms of \(H_i\), though we did not
do this in this case. We filter the hypotheses found by \LGEM{} to
just those that might be testable via an experiment, and used flux
balance analysis to select plausible candidates. Hypotheses that
correct errors with minimal effect on the current model (produce as
few as possible new metabolites) are selected. This is in accordance
with the principle of parsimony as discussed in
Section~\ref{parsimony}. For more details of these processes, along
with further examples of deductions and abduction of hypotheses can
be found in~\cite{gowerLGEMFirstOrderLogic2023}.

The hypotheses that \LGEM{} generates may well be very close together in
the experimental space. The interactions between the components of such a
complex system mean that a small difference in input could have a large
effect on the outcome of the experiment. This is one of the benefits of
using a robot scientist to execute the experiments. As we mentioned in
Section~\ref{the-philosophy-of-robot-scientists} we can achieve higher
precision with robots and also log more data that could help explain
systematic errors. Yeast cultivations typically stretch over a few days,
so by using robots, tiredness becomes less of an issue for the human
scientists involved in the project. Genesis is also equipped with a
microformulator allowing for fine adjustment of input media for the
cultures, to a degree that would be onerous to replicate with handheld
pipettes or even a traditional liquid handling robot.

\subsection{Interpretation from ML
paradigms}\label{interpretation-from-ml-paradigms}

As we can see from the stated goal in the application domain, the
criteria for success of the resulting model is to measure its predictive
performance against empirical results. Mapping between input and output
pairs is non-trivial in a lot of cases. The information obtained through
experimental measurements does not always align directly with what \LGEM{}
can predict, particularly with relation to current metabolomics methods.
However, despite the imperfect mapping of predicted to observed outputs,
this follows the supervised learning paradigm.

Models like \LGEM{} provide plausible hypothesis candidates for
testing and feeding back to the discovery loop, which is aligned with
the active learning paradigm and targeted exploration. When exploring
the experimental space, due to the complexity of the system and the
stochastic factors, certain phenomena might require many experiments
that are very close together in the experimental space. This requires
precision, and Genesis has been designed to provide this precision.
The other selection criterion was to increase diversity in the
dataset. A general purpose microformulator and modular design were
deliberately chosen to increase options for experimentation.

\section{Conclusions and Future
Directions}\label{conclusions-and-future-directions}

In this chapter, we covered the core concepts of the scientific method as
they relate to robot scientists. We provided some tools for the analysis
of robot scientist projects, namely to consider them through an active
learning paradigm, and to map the values of scientific models onto
techniques from machine learning. We discussed an example application
domain, systems biology and a next-generation robot scientist, Genesis.
We concluded by showing \LGEM{}, a first order logic (FOL) model for
Genesis.

One motivation for choosing FOL to model yeast metabolism was grounded in
epistemic parsimony; FOL frameworks are easily extensible. There are many
phenomena that could be relevant for experimentation, that \LGEM{} does not
currently model. But the logical framework can be updated without
affecting the underlying infrastructure. We saw from the results on the
single-gene essentiality prediction that there is a gap in explanation.
It is necessary to to accede a more complex model, and the background knowledge
suggests that we should incorporate new mechanisms such as gene
regulation to capture some of the higher order behaviours of the system.

Our next steps will be working toward full integration of the
computational models with the Genesis hardware, relying on controlled
vocabularies~\cite{rederGenesisDBDatabaseAutonomous2023} to specify
experiments and results so that we can map inputs to outputs for the
machine learning algorithms. These are necessary steps so that we can
employ computational models like \LGEM{} with Genesis for closed-loop
experimentation.

\begin{credits}
    \subsubsection{\ackname}
    This work was partially supported by the Wallenberg AI, Autonomous
    Systems and Software Program (WASP) funded by the Alice Wallenberg
    Foundation. Funding was also provided by the Chalmers AI Research Centre
    and the UK Engineering and Physical Sciences Research Council (EPSRC)
    grant nos: EP/R022925/2 and EP/W004801/1, as well as the Swedish Research
    Council Formas (2020-01690).
\end{credits}

\bibliographystyle{splncs04}
\bibliography{references}

\end{document}